\documentclass[10pt,twocolumn,letterpaper]{article}

\usepackage{iccv}
\usepackage{times}
\usepackage{epsfig}
\usepackage{graphicx}
\usepackage{amsmath}
\usepackage{amssymb}
\usepackage{svg}

\usepackage{float}

\newcommand{\rebuttal}[1]{#1}

\newcommand{\Real}{{\rm I\!R}}
\usepackage{tikz}
\usepackage{pgfplots}
\usepackage{multirow, tabularx}
\usepackage{mathtools}
\usepackage{svg}

\usepackage{cancel,amsmath}
\usepackage{booktabs}
\usepackage{tikz}
\usepackage{graphicx}
\usepackage[caption=false]{subfig}

\usepackage{colortbl}
\usepackage{xcolor}
\definecolor{redx}{rgb}{1.0, 0.9, 0.9}
\definecolor{bluex}{rgb}{0.94, 0.97, 1.0}

\usepackage[backend=bibtex,style=numeric-comp]{biblatex}
\bibliography{references, extra_references}
\usetikzlibrary{calc}

\usepackage{footmisc} 
\usepackage{adjustbox}	
\usepackage{stackengine}

\definecolor{blue_eqn}{RGB}{56,100,252}
\definecolor{red_eqn}{RGB}{217,88,71}

\usepackage[pagebackref=false,breaklinks=true,letterpaper=true,colorlinks,bookmarks=false]{hyperref}

\iccvfinalcopy 

\def\iccvPaperID{8} 

\ificcvfinal\fi

\begin{document}

\title{Reconstructing Pruned Filters using Cheap Spatial Transformations}

\author{Roy Miles\\
Imperial College London\\
{\tt\small r.miles18@imperial.ac.uk}
\and
Krystian Mikolajczyk\\
Imperial College London\\
{\tt\small k.mikolajczyk@imperial.ac.uk}
}


\maketitle

\begin{abstract}
We present an efficient alternative to the convolutional layer using cheap spatial transformations. This construction exploits an inherent spatial redundancy of the learned convolutional filters to enable a much greater parameter efficiency, while maintaining the top-end accuracy of their dense counter-parts. Training these networks is modelled as a generalised pruning problem, whereby the pruned filters are replaced with cheap transformations from the set of non-pruned filters. We provide an efficient implementation of the proposed layer, followed by two natural extensions to avoid excessive feature compression and to improve the expressivity of the transformed features. We show that these networks can achieve comparable or improved performance to state-of-the-art pruning models across both the CIFAR-10 and ImageNet-1K datasets. 
\end{abstract}

\section{Introduction}
\label{sec:intro}
\begin{figure*}
    \centering
    \includegraphics[width=0.7\linewidth]{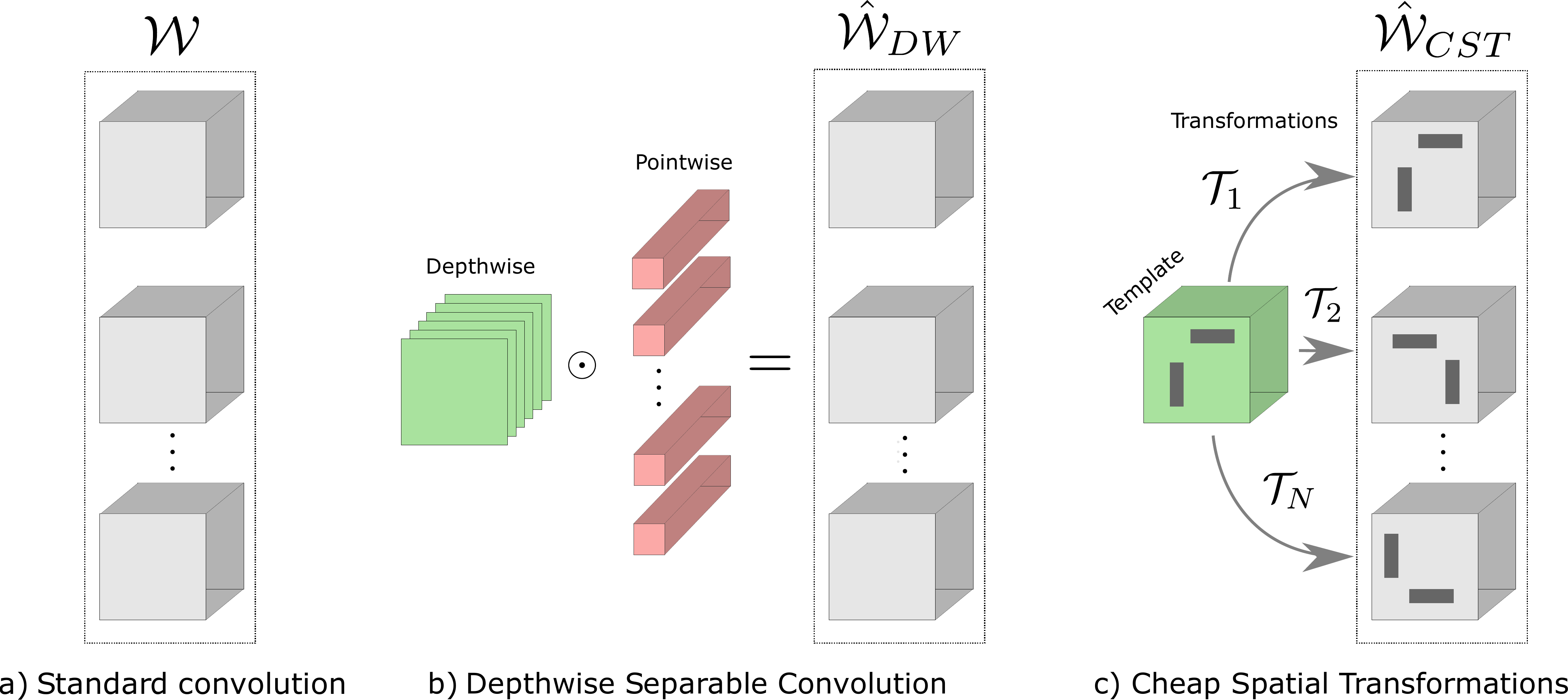}
    \vspace{1em}
    \caption{\textbf{Constructing more convolutional filters using cheap spatial transformations.} (a) Original convolutional layer. (b) Depth-wise separable layers, which fully decouple the spatial and depthwise aggregation of features. (c) Proposed layer expressed new filters as spatial transformations of a smaller set of templates.}
    \label{fig:overview}
\end{figure*}

Convolutional neural networks (CNNs) have achieved state-of-the-art results across a range of computer vision tasks~\cite{Simonyan2015VeryRecognition, Detone2018SuperPoint:Description}. Despite their success, these models are typically far too large and computationally expensive for their deployment on resource constrained devices, such as mobile phones or edge devices. Recent work on compressing CNNs \cite{Li2017PruningConvnets, Zhuang2018Discrimination-awareNetworks, Wang2019COP:Pruning, Lin2019FilterPruning, Ding2019ApproximatedOptimization, Qin2019CAPTOR:Applications, He2018FilterAcceleration, He2022FilterNetworks}, have exploited the inherent weight redundancy using structured pruning. This approach provides a way to reduce the size of a network without relying on sparse software libraries or hardware accelerators. Most of these methods involve ranking the importance of filters and then removing those that fall below a specific threshold. However, it is important to note that when the pruning rates are high, some of these pruned filters can still contribute in retaining the top-end accuracy. To address this limitation, we propose a cheap decomposition of the convolutional layer where the pruned filters are reconstructed using cheap spatial transformations of the non-pruned filters, which we call templates. We propose an approach to transfer an existing CNN to this efficient architecture through a generalised pruning pipeline. This methodology can be considered a natural extension of pruning, but instead of zeroing out the pruned filters, we are replacing them with the cheap template transformation. This work can be related to group equivariant convolutional networks~\cite{Cohen2016GroupNetworks}, which consider the hand-crafted construction of filters using a pre-defined group to learn equivariant features. In contrast to this work, we jointly learn both the transformations and templates with the alternative objective of training small and efficient CNNs. Our contributions can be summarised as follows:
\begin{itemize}
    \item We propose a novel approach to construct expressive convolutional filters from cheap spatial transformations using a set of filter templates.
    \item We model the training as a generalised pruning problem with a simple magnitude based saliency measure. 
    \item We introduce a grouped extension to mitigates excessive feature compression.
    \item Our results show competitive performance over state-of-the-art pruning methods on both the CIFAR-10 and ImageNet-1K datasets.
\end{itemize}

\subsection{Related work}
The most relevant work can be divided into pruning, low-rank decomposition, and knowledge distillation.

\textbf{Pruning} explicitly exploits the inherent parameter redundancy by removing individual weight entries or entire filters that have the least contribution to the  performance on a given task. This was first introduced in~\cite{Lecun1990OptimalDamage, Hassibi1993SecondSurgeon} using the Hessian of the loss to derive a saliency measure for the individual weights. \rebuttal{SNIP~\cite{Lee2019SnIP:Sensitivity} proposed to prune weights using the connection sensitivity between individual neurons. Subsequent work propose a sparse neuron skip layer~\cite{Subramaniam2020N2NSkip:Connections} to achieve fast training convergence and a high connectivity between layers.}
Cheap heuristic measures have also been used, such as the magnitude~\cite{Han2015DeepCoding, Li2017PruningConvnets}, geometric median~\cite{He2018FilterAcceleration}, or average percentage of zeros~\cite{Hu2016NetworkArchitectures}. Although some of these unstructured pruning methods are able to achieve significant model size compression, the theoretical reduction in floating-point operations (FLOPs) does not translate to the same practical improvements without the use of dedicated sparse hardware and software libraries. This has led to the more widespread adoption of structured pruning approaches, which focus on removing entire filters. \cite{Zhuang2018Discrimination-awareNetworks} introduced additional loss terms to select the channels with the highest discriminative power, while \cite{Yu2018NISP:Propagation} proposed to prune in accordance with a neural importance score. DMCP~\cite{Guo2020DMCP:Networks} models the pruning operation as a differentiable markov chain, where compression is achieved through a sparsity inducing prior. Similarly,  \cite{Baalen2020BayesianPruning,Zhao2019VariationalPruning} model pruning in the probabilistic setting using both hierarchical and sparsity inducing priors. Unlike these prior works, we propose to reconstruct the pruned filters using cheap transformations. These reconstructed filters are shown to be expressive and learn diverse features, thus mitigating the need for any sophisticated pruning strategies.

\textbf{Low-rank decomposition} is concerned with compactly representing a high-dimensional tensor, such as the convolutional weights, as linear compositions of much smaller, lower-dimensional, tensors, called factors. Any linear operations that are then parameterised by these weights can be expressed using cheaper operations with these factors, which can lead to a reduction in the computational complexity.
Depthwise separable convolutions split the standard convolution into two stages, the first extracts the local spatial features in the input, while the second aggregates these features across channels.
They were originally proposed in Xception~\cite{Chollet2017Xception:Convolutions} but have since been adopted in the design of a range of efficient models~\cite{Chen2018AnMobileNet, Fox2018MobileNetV2:Bottlenecks, Howard2019SearchingMobileNetV3, Tan2018MnasNet:Mobile}. This has led to the development of optimized GPU kernels that bridge the gap between the theoretical FLOP improvements and the practical on-device latency. 
Both CP-decomposition~\cite{Hitchcock2015TheProducts} and Tucker decomposition~\cite{Tucker1966SomeAnalysis} have also been used to construct or compress pre-trained models~\cite{Jaderberg2014SpeedingExpansions, Kim2015CompressionApplications, Miles2021CompressionApplications}. Another line of work has explored the use of tensor networks as a mathematical framework for generalising tensor decomposition in the context of deep learning~\cite{Hayashi2019Einconv:Networks, Wang2018WideNets}. Ghost modules~\cite{Han2019GhostNet:Operations} use depthwise convolutions to construct more features, leading to improved capacity at a much smaller overhead. Our proposed layer can be seen as an alternative parameterisation of the convolutional weights which can be naturally pruned using a generalised pruning pipeline.

\textbf{Knowledge distillation} attempts to transfer the knowledge of a large pre-trained model (teacher) to a much smaller compressed model (student). This was originally introduced in the context of image classification~\cite{Hinton2015DistillingNetwork}, whereby the soft predictions of the teacher can act as pseudo ground truth labels for the student. This methodology enables the student model to more easily learn the correlations between classes which are not available through the one-hot encoded ground truth labels. Hinted losses further provide knowledge distillation for the intermediate representation~\cite{Romero2015FitNets:Nets} and can be modelled as reconstruction L2 loss terms in the same space or in a projected feature space~\cite{Yim2017ALearning, Qian2020EfficientDistillation, Miles2023MobileVOS:Distillation, Chen2022ImprovedEnsemble, Miles2023ADistillationv2, Miles2022InformationDistillation}. Weight sharing and jointly training models at different widths/pruning-rates has also been shown to provide implicit knowledge distillation to the smallest models~\cite{Yu2018SlimmableNetworks, Yu2019UniversallyTechniques}. In general, our proposed method is orthogonal to knowledge distillation - its adoption can be employed in addition to further improve performance.


\section{Method}
\label{sec:method}
In this section we propose a novel decomposition of the convolutional layer. We do this be expressing the convolutional filters as spatial transformations of a compact set of template filters. These templates are obtained through a well-established pruning procedure, ensuring discriminative features. Subsequently, we present an algorithmically equivalent derivation of this layer that has much fewer floating-point operations (FLOPs). Moreover, we extend our approach naturally by introducing a group extension, which enhances the connectivity between layers. This extension enables an improved channel connectivity, fostering more robust and informative feature propagation.

\subsection{Constructing diverse convolutional filters.}
\label{sec:decomposition}
Let $\mathcal{W} = \{\mathcal{W}_n \in \Real^{K \times K \times C}\}_{n=1}^N$ describe the set of filters for a given convolutional layer with an input depth $C$, output depth $N$, and a receptive field size of $K \times K$. Our method is based on an assumption that a large subset of these filters can be faithfully approximated as spatial transformations of a much smaller set of filters, which we call templates $\mathcal{B}$. For simplicity, and without loss in generality, consider the scalar transformations, which can be implemented as cheap element-wise products between the spatial entries of the templates. Thus, for a $K \times K \times C$ template, each transformation can be parameterised using $K \times K$ learnable weights. Consider the case with $N$ templates and $N$ output feature maps. The proposed decomposition is given as follows:

\begin{align}
    \mathcal{Y}_{h, w, n} &= \sum_{k_w, k_h}^K\sum_{i}^C \mathcal{X}_{h', w', i} \cdot \mathcal{W}_{k_h, k_w, c, n} \\
    &\approx \sum_{k_w, k_h}^K\sum_{i}^C \mathcal{X}_{h', w', i} \cdot \mathcal{B}_{k_h, k_w, i, n} \cdot \mathcal{T}_{k_h, k_w, n} \label{eq:basis} \\
    h' = (h - &1)s + k_h - p, \;\;\;\; w' = (w - 1)s + k_w - p \nonumber
\end{align}

where $s$ is the stride and $p$ is zero-padding. The general formulation using affine transformations is illustrated in figure \ref{fig:overview}.
Model compression can be achieved when the number of basis filters $M$ is less than the number of output feature maps $N$. This is realised through pruning, which is discussed in second \ref{sec:pruning}. In this case, the templates are then re-used to compute more filters using different transformations.

The choice of mapping from which template to which output feature map is not critical, as long as it is fixed after the pruning stage to enable fine-tuning. For our experiments, we set this mapping to be $i = j \; mod \; M$, where the $i$th output feature map is allocated the $j$th template, and where $M$ is the total number of templates. This choice of mapping ensures that all templates are uniformly used, thus enabling a diverse set of transformed filters. The spatial transformations are then jointly learned alongside the templates.

\begin{figure}[t]
\centering
\subfloat
{
    \includegraphics[width=0.80\linewidth]{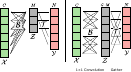}
}
\caption{\textbf{Too few templates can overly compress the input features.} We propose to introduce a group parameter to naturally balance the expressiveness of the both the template and transformation stages.}
\vspace{-1em}
\label{fig:channel_connectivity}
\end{figure}

\begin{figure*}
\centering
\subfloat
{
    \adjustbox{width=.25\linewidth}{
    \includegraphics{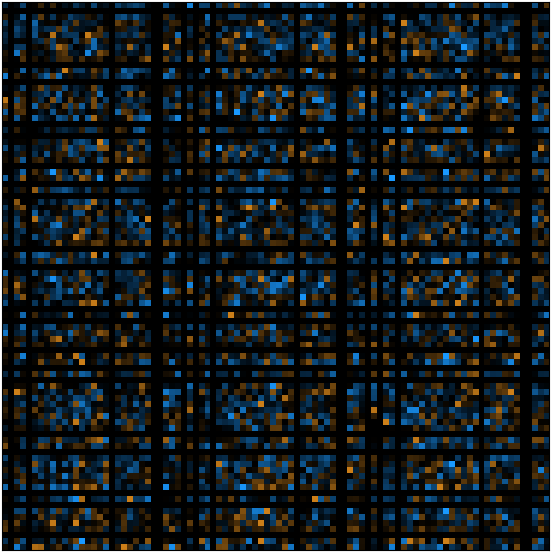}
    }
    \adjustbox{width=.25\linewidth}{
    \includegraphics{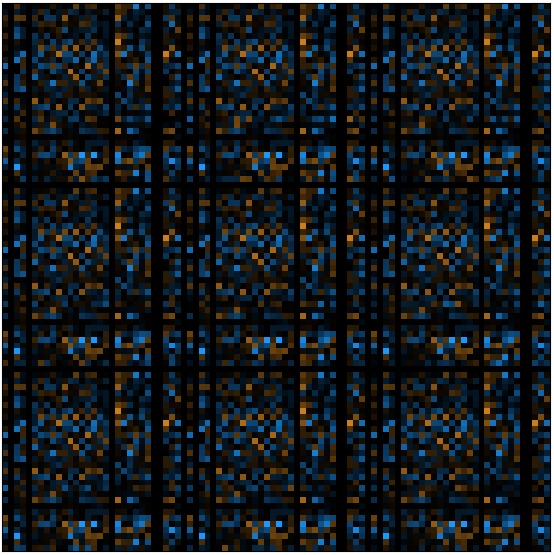}
    }
    \adjustbox{width=.25\linewidth}{
    \includegraphics{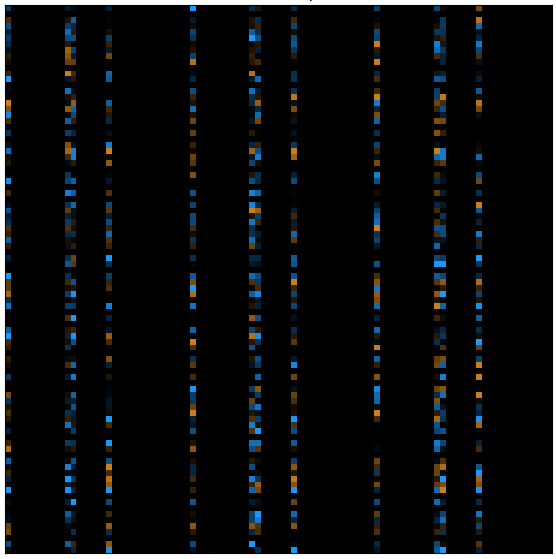}
    }
}
\caption{\textbf{Visualising learned transformations for a given layer.} (left) original learned filters. (middle) expressive filters using affine template transformations. (right) pruning filters. Both the vanilla pruned layer and the decomposed layers use a pruning rate of 0.7. Each column represents a filter for a given output channel and black pixels represent zero entries.}
\label{fig:visualising_learned_transformations}
\end{figure*}

\subsection{Using pruning to select the filter templates.}
\label{sec:pruning}
\begin{figure}[h]
\centering
\adjustbox{width=.65\linewidth}{
\includegraphics{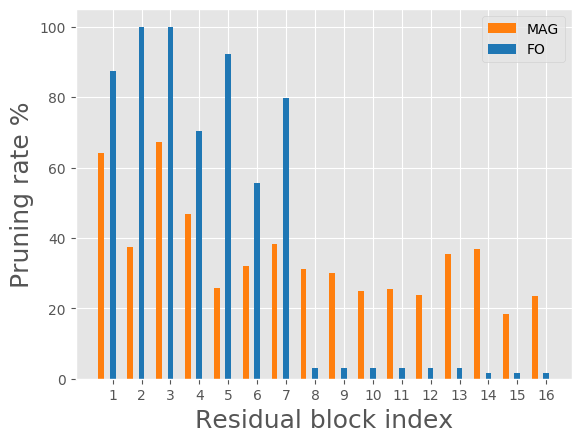}
}
\caption{\textbf{Comparing the pruning rates of each layer using different saliency measures.} We see that a magnitude based (MAG) criterion exhibits a more uniform pruning rate than gradient based measures (FO).}
\label{fig:pruning_allocation}
\end{figure}

The pruning literature has proposed increasingly sophisticated pruning heuristics and training pipelines. Examples of such including layer-wise pruning strategies~\cite{dong2017learning} and gradient-based saliency measures~\cite{molchanov2019importance}, which incur additional hyperparameters and increased computational costs. In favour of simplicity, and to demonstrate the generalisability of our decomposition, we propose to use a very simple magnitude-based criterion to rank the importance of filters for selecting the set of templates. We observe that this choice of saliency measure naturally leads to a uniform pruning strategy across all layers in the network (figure \ref{fig:pruning_allocation}), which reduces excessive feature compression for any given layer (see section \ref{sec:channel_connectivity}).

\begin{figure}[H]
\centering
\subfloat
{
    \adjustbox{width=.45\linewidth}{
    \includegraphics{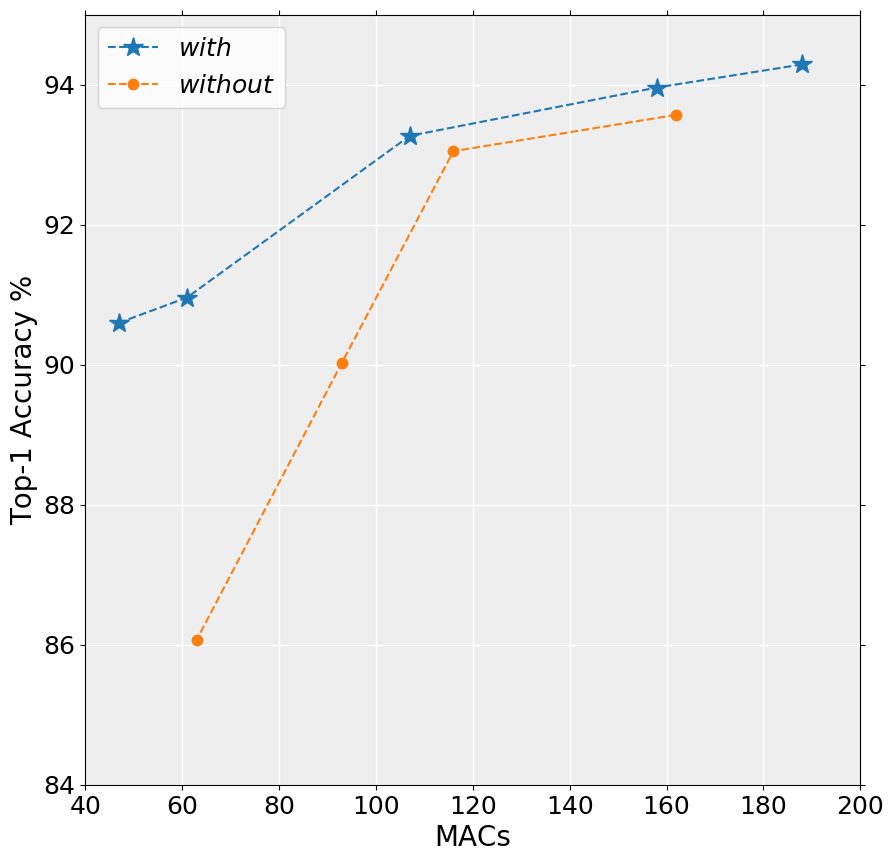}
    }
    \adjustbox{width=.45\linewidth}{
    \includegraphics{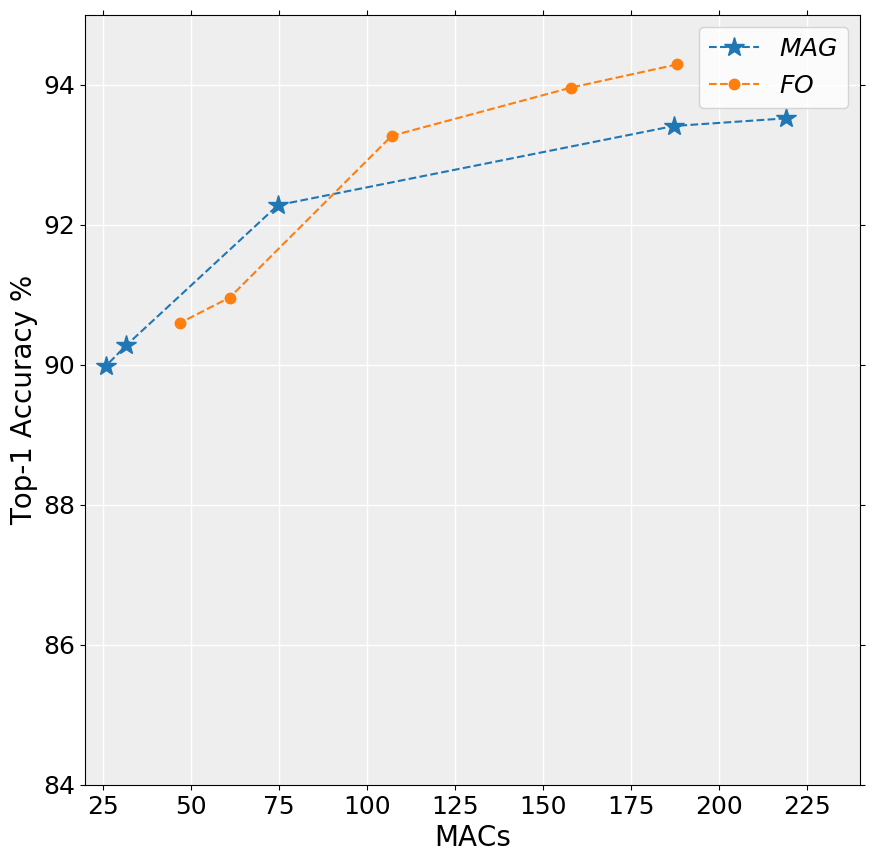}
    }
}
\caption{\textbf{Comparison with a pruned baseline and the importance of which pruning heuristic.} (left) Performance of a VGG16 network trained using affine (with) and scalar (without) spatial transformations with the same training methodology. (right) show the performance trade-off at different pruning rates using both a magnitude-based and a gradient-based saliency.}
\label{fig:pruning_heuristic}
\end{figure}

We provide an ablation on the importance on the choice of saliency measure in figure \ref{fig:pruning_heuristic} (right). In this ablation we compare the performance using two different measures, namely magnitude based and gradient based~\cite{molchanov2019importance}. Although in some cases the gradient based measure does lead to better performance, which is attributed to a more discriminative selection of templates, it does come at an increased computational overhead. In favour of simplicity, and to demonstrate the robustness to the choice of templates, we use a magnitude based measure throughout. In fact, for the ImageNet-1K experiments we extend this hypothesis and use a randomly initialised network to begin with, rather than from a pre-trained network - as is more commonly used in the pruning literature.

\subsection{Performance and efficient implementations.}
\label{sec:efficient_implementation}

Computing the output features using the transformed filters and then using a standard convolution would not lead to any reduction in FLOPs. To address this, we propose to decompose the convolutional layer into two stages. The first computes the template \textit{features} $\mathcal{Z}$ using the template filters $\mathcal{B}$, while the second projects these features to a different space using the spatial transformations $\mathcal{T}$. The output features are then the union of the original template features (identity transformations) and the transformed features. This two stage implementation is algorithmically equivalent to first constructing the filters and then performing a convolution, and its derivation is given as follows:
\vspace{-1em}
\begin{align}
     \mathcal{Y}_{h, w, n} 
    &\approx \sum_{k_w, k_h}^K\sum_{i}^C \mathcal{X}_{h', w', i} \cdot \mathcal{B}_{k_h, k_w, i, n} \cdot \mathcal{T}_{k_h, k_w, n} \\
    &= \sum_{k_w, k_h}^K \mathcal{T}_{k_h, k_w, n}  \underbrace{\left(\sum_{i}^C \mathcal{X}_{h', w', i} \cdot \mathcal{B}_{k_h, k_w, i, n}\right)}_{\mathcal{Z}}     \label{eq:basis2}
\end{align}
Computing $\mathcal{Z}$ can be achieved using a pointwise convolution, which translates to an optimised general matrix multiply primitive. The second stage, which consists of projecting $\mathcal{Z}$ to the output space, reduces to a series of gather operations and multiplications, which will implement the spatial transformations. Both of these operations can be trivially implemented in most deep learning frameworks. 

\subsection{Channel connectivity and feature compression.}
\label{sec:channel_connectivity}
By design, the proposed decomposition preserves the same number of input and output channels as the original convolution. This means that all the pruned filters are being reconstructed using some cheap and learned template transformation. The consequence of this design is that at high pruning rates there will be a significant bottleneck in the latent space $\mathcal{Z}$ (see figure \ref{fig:channel_connectivity}). 
This bottleneck can result in significant feature compression that can degrade the downstream performance and discriminative power of representations.
We could address this problem by simply increasing the number of templates per layer, but this would incur a significant overhead in terms of both parameters and FLOPs. Instead, we propose to introduce a grouped extension that can naturally scale the dimensionality of the latent space $\mathcal{Z}$ with a minimal computational and parameter overhead. To do this we replace the {\em pointwise convolution} in the two stage processing with a {\em grouped pointwise convolution}~\cite{Krizhevsky2012ImageNetNetworks}, which has an efficient implementation in most deep learning frameworks. This transformation then translates to the sum of $G$ transformations applied to feature maps from the $G$ distinct groups. Doing so in this way enables cross-group information flow without the need for any channel shuffles~\cite{Zhang2018ShuffleNet:Devices}.
Figure \ref{fig:channel_connectivity} graphically demonstrates this grouped extension. On the left is the original case, whereby $G = 1$. At this pruning rate, there is a very large compression of features. Increasing the groups to 2 (as shown on the right) provides a natural scheme for increasing the depth of $\mathcal{Z}$ without incurring any significant computational overhead. The results of the $G$ different transformations across groups are then added to form each of the $N$ output channels.

\begin{figure}[H]
\centering
\subfloat
{
    \adjustbox{width=.45\linewidth}{
    \includegraphics{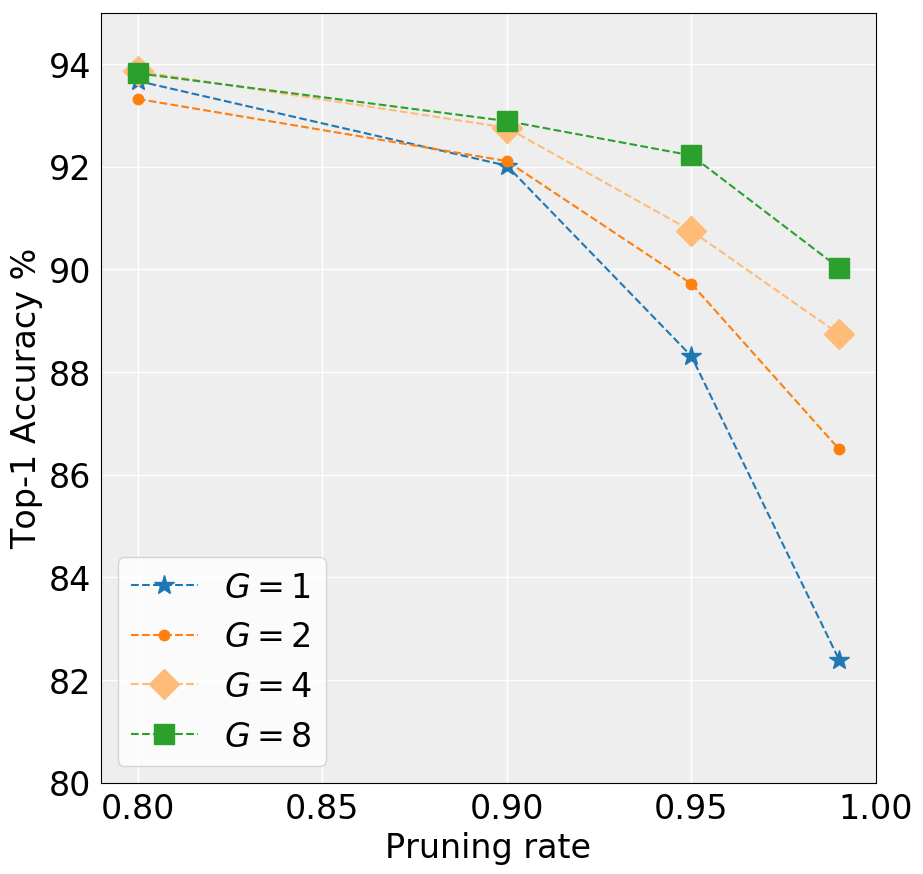}
    }
    \adjustbox{width=.45\linewidth}{
    \includegraphics{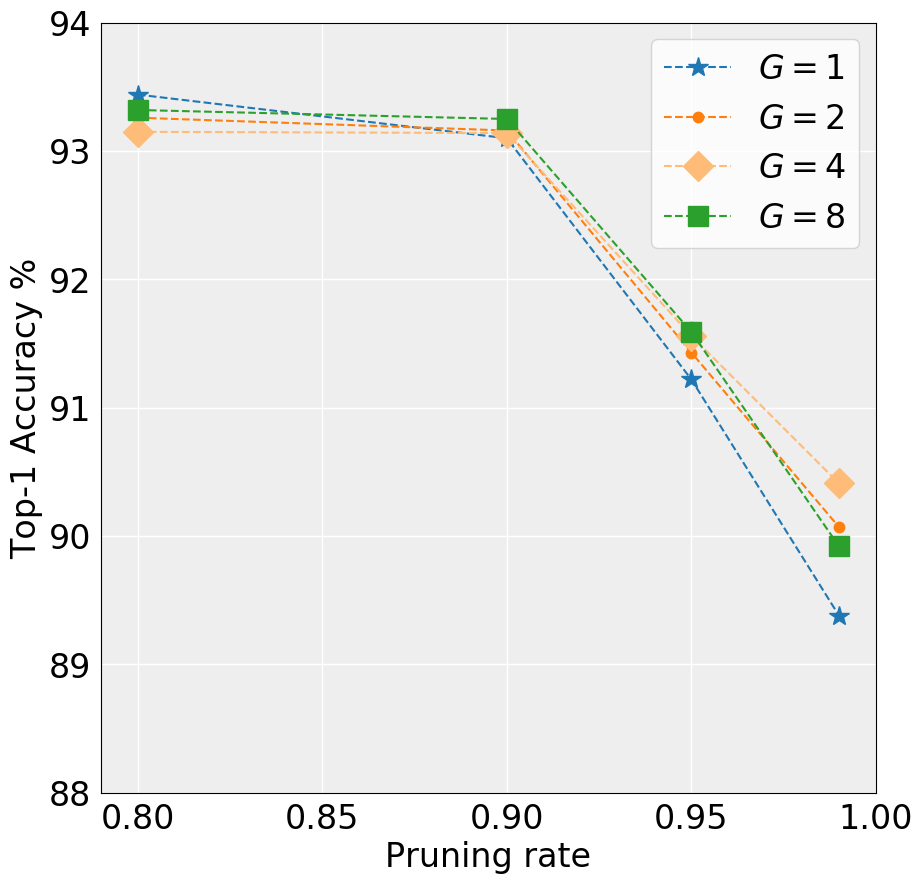}
    }
}
\caption{\textbf{Importance of channel connectivity and the group extension.} (left) Using more groups enables a smaller bottleneck ratio, which improves the top-end accuracy. (right) shows that increasing the minimum number of templates also reduces feature compression but at a much larger overall cost.}
\label{fig:group_ablation}
\end{figure}

In the limiting case where $G = M$, the channel connectivity pattern is very similar to that of depthwise-separable convolutions since there will be a one-to-one mapping between channels in the first stage, while the second stage will be fully connected. However, there is still a significant distinction between the two - our proposed decomposition enables spatial aggregation of features in both stages. \\

Figure \ref{fig:channel_connectivity} demonstrates the importance of this grouped extension at high pruning rates. We find that although simply increasing the minimum number of templates in each layer does implicitly address this feature compression problem, it comes at a much larger overall cost. In general, $G$ can be tuned depending on the target pruning rate.

\subsection{Computational cost and parameter efficiency.}
\label{sec:computational_and_parameter_cost}

The standard convolutional layer has the computational cost of the order of $H \cdot W \cdot K^2 \cdot C \cdot N$, whereas the cost of the proposed decomposition is given by:

\begin{align}
    FLOPS = & \; H \cdot W \cdot K^2 \cdot \frac{C}{\cancel{G}} \cdot M \cdot \cancel{G} + \\ \nonumber
    &H \cdot W \cdot K^2 \cdot G \cdot (N - M)
\end{align}

Where $M$ indicates the number of templates and $G$ is the number of groups. The reduction in computation ($FLOPS\downarrow$) is subsequently given by:

\begin{align}
    FLOPS\downarrow &= \frac{HWK^2 \cdot C \cdot M + HWK^2 \cdot G \cdot (N -M )}{HWK^2 \cdot C \cdot N} \\
    &= {\color{blue_eqn} \frac{M}{N} + \frac{G}{C}} - {\color{red_eqn} \frac{GM}{CN}}
    \label{eqn:flop_reduction}
\end{align}

We prune the set of templates such that $M \ll N$ and we set $G \ll C$ to yield a reduction in FLOPs. We further improve the bottleneck problem by using $G\cdot M$ templates applied to $C/G$ channels that are efficiently implemented with grouped convolutions. Finally, we ensure cross-group information flow by increasing the number of cheap spatial transformations that are then applied cross group.

From a similar view, we can also derive the reduction in parameters, where the number of parameters for a convolutional layer is given by:

\begin{align}
    PARAMS = K^2 \cdot C \cdot N
    \label{eqn:params_conv}
\end{align}

and our proposed decomposition has a parameter count given by:

\begin{align}
    PARAMS = K^2 \cdot \frac{C}{G} \cdot M + K^2 \cdot G \cdot (N - M)
    \label{eqn:params_decomp}
\end{align}

Not the subtraction is because we use $M$ identity transformation, while the rest of the output features are computed using cheap spatial transformations. Using both \ref{eqn:params_conv} and \ref{eqn:params_decomp}, we can derive the reduction in parameters ($PARAMS\downarrow$), which ends up being identical to equation \ref{eqn:flop_reduction}.

\begin{align}
    PARAMS\downarrow &= \frac{K^2 \cdot \frac{C}{G} \cdot M + K^2 \cdot G \cdot (N - M)}{K^2 \cdot C \cdot N} \\
    &= {\color{blue_eqn} \frac{M}{GN} + \frac{G}{C} } - {\color{red_eqn} \frac{GM}{CN} } 
    \label{eqn:param_reduction}
\end{align}

When using more general and expressive spatial transformations, such as $GL(3)$ or $SO(3)$, the second stage can instead be implemented using a bilinear sampling of neighbouring spatial pixels in $\mathcal{Z}$. These transformations will be parameterised using a $2 \times 3$ matrix and result in the number of floating point operations being increased to $4$ per spatial location. This increase in FLOPs and the number of parameters is often small, but it enables a significant increase in the expressiveness of transformations. However, in general, we observe that a learned scalar transformations can still yield a strong accuracy v.s. performance trade-off (see figure \ref{fig:channel_connectivity} left and table \ref{table:ablation_transformations}). We wish to highlight that in the case of these more general spatial transformations, the parameter reduction equation \ref{eqn:param_reduction} and flop reduction equation \ref{eqn:flop_reduction} will differ.

\section{Experiments}
\label{sec:experiments}
\begin{table*}
    \centering
    \adjustbox{width=.9\linewidth}{
    \begin{tabular}{clccccc}
        \toprule
        
        Model & Method & Baseline Acc. (\%) & Acc. (\%) & Acc. Drop (\%) & FLOPs $\downarrow$ (\%) & Parameters $\downarrow$ (\%) \\
        \midrule
        \multirow{8}{*}{VGG16} & Hinge~\cite{Li2020GroupCompression} & 93.59 & 94.02 & -0.43 & 39.07 & 19.95 \\
        & \cellcolor{redx}\textbf{Ours} & \cellcolor{redx}\textbf{93.26} & \cellcolor{redx}\textbf{93.92} & \cellcolor{redx}\textbf{-0.66} & \cellcolor{redx}\textbf{45.56} & \cellcolor{redx}\textbf{56.77} \\
        & NSPPR~\cite{Zhuang2022Neuron-levelRegularizer} & 93.88 & 93.92 & -0.04 & 54.00 & - \\
        & AOFP~\cite{Ding2019ApproximatedOptimization} & 93.38 & 93.84 & -0.46 & 60.17 & - \\
        & DLRFC~\cite{He2022FilterNetworks} & 93.25 & 93.93 & -0.68 & 61.23 & 92.86 \\
        & DPFPS~\cite{Ruan2021DPFPS:Scratch} & 93.85 & 93.67 & 0.18 & 70.85 & 93.92 \\
        & \cellcolor{redx}\textbf{Ours} & \cellcolor{redx}\textbf{93.26} & \cellcolor{redx}\textbf{93.62} & \cellcolor{redx}\textbf{-0.36} & \cellcolor{redx}\textbf{61.38} & \cellcolor{redx}\textbf{81.15} \\
        & ABC~\cite{Lin2020ChannelSearch} & 93.02 & 93.08 & -0.06 & 73.68 & 88.68 \\
        & HRank~\cite{Lin2020HRank:Map} & 93.96 & 91.23 & 2.73 & 76.50 & 92.00 \\
        & AOFP~\cite{Ding2019ApproximatedOptimization} & 93.38 & 93.28 & 0.10 & 75.27 & - \\
        & \cellcolor{redx}\textbf{Ours} & \cellcolor{redx}\textbf{93.26} & \cellcolor{redx}\textbf{92.74} & \cellcolor{redx}\textbf{0.52} & \cellcolor{redx}\textbf{80.89} & \cellcolor{redx}\textbf{95.26} \\
        \midrule
        & NISP~\cite{Yu2018NISP:Propagation} & 93.04 & 93.01 & 0.03 & 43.60 & 42.60 \\
        \multirow{8}{*}{ResNet56} & \cellcolor{redx}\textbf{Ours} & \cellcolor{redx}\textbf{93.60} & \cellcolor{redx}\textbf{94.18} & \cellcolor{redx}\textbf{-0.58} & \cellcolor{redx}\textbf{35.91} & \cellcolor{redx}\textbf{51.64} \\
        & FPGM~\cite{He2018FilterAcceleration} & 93.59 & 93.49 & 0.10 & 53.00 & - \\
        & NSPPR~\cite{Zhuang2022Neuron-levelRegularizer} & 93.83 & 93.84 & -0.03 & 47.00 & - \\
        & ABC~\cite{Lin2020ChannelSearch} & 93.26 & 93.23 & 0.03 & 54.13 & 54.20 \\
        & SRR-GR~\cite{Wang2019COP:Pruning} & 93.38 & 93.75 & -0.37 & 53.80 & - \\
        & \cellcolor{redx}\textbf{Ours} & \cellcolor{redx}\textbf{93.60} & \cellcolor{redx}\textbf{93.35} & \cellcolor{redx}\textbf{0.25} & \cellcolor{redx}\textbf{48.27} & \cellcolor{redx}\textbf{71.24} \\
        & DPFPS~\cite{Ruan2021DPFPS:Scratch} & 93.81 & 93.20 & 0.61 & 52.86 & 46.84 \\
        & \cellcolor{redx}\textbf{Ours} & \cellcolor{redx}\textbf{93.60} & \cellcolor{redx}\textbf{92.81} & \cellcolor{redx}\textbf{0.79} & \cellcolor{redx}\textbf{56.36} & \cellcolor{redx}\textbf{80.05} \\

        \bottomrule
    \end{tabular}%
    }
    \vspace{0.5em}
    \caption{\textbf{Comparison to other pruning methods on CIFAR10.} Each model is trained using a magnitude based measure for selecting the filter templates, number of groups = 2, and with a minimum of 8 template filters per layer to avoid catastrophic pruning.}
    \label{table:cifar10_results}
\end{table*}

In this section we evaluate our approach on the CIFAR and ImageNet datasets. The models are compared through the number of parameters, the number of floating point operations, followed by the top-1 classification accuracy. All of these models are trained on a single NVIDIA RTX 2080Ti GPU using either stochastic gradient descent (CIFAR10) or AdamW (ImageNet-1K). We set the minimum number of templates for each layer to be 8 for CIFAR10 and 32 for ImageNet-1K. Finally, we use the number of groups in each layer to be 2 for all the main benchmark experiments

\subsection{Experimental results on CIFAR-10}
\label{sec:cifar_experiments}
The CIFAR10 dataset \cite{Krizhevsky2009LearningImages} consist of 60K $32 \times 32$ RGB images across 10 classes and with a 5:1 training/testing split. The chosen VGG16~\cite{Simonyan2015VeryRecognition} architecture is modified for this dataset with batch normalisation layers after each convolution block and by reducing the number of classification layers to one. During training, we augment the datasets using random horizontal flips, random $32\times32$ crops, and random rotations. The baseline architectures are trained for 300 epochs with a step learning rate decay and we use a simple magnitude based criterion for ordering and selecting the most important filters to form the set of templates. This selection is in conjunction with a simple linear pruning schedule that spans the first 40 epochs of training. We highlight that this choice of pruning schedule is in contrast to most of the other pruning methods~\cite{Zhuang2018Discrimination-awareNetworks, Luo2017ThiNet:Compression}, which can adopt much longer pruning stages and introduce additional layer-by-layer stopping conditions. The results are shown in table \ref{table:cifar10_results} and show comparable or improved performance to the much more sophisticated pruning strategies across a wide range of compression ratios.

\subsection{Experimental results on ImageNet-1K}
\label{sec:imagenet_experiments}
For experiments on ImageNet-1K we use the ResNet-50 architecture and the same magnitude based saliency measure described in section \ref{sec:cifar_experiments}. In general, magnitude pruning was empirically shown to provide a more uniform pruning rate across all of the residual blocks, which is important to avoid overly compressing intermediate features. We set the minimum number of templates for each layer to be 8 and the number of groups in each layer to be 2. The model is trained for a 300 epochs with a linear learning rate decay every 25 epochs. Finally, we use MixUp and CutMix augmentations with $\alpha$ set to $0.1$ and $1.0$ respectively.

The ImageNet results are shown in table \ref{table:imagenet_results} and show comparable performance to state-of-the-art pruning methods without the need for extensive pruning and fine-tuning pipelines. To further demonstrate the robustness of this decomposition to the choice of templates and the effectiveness of joint template/transformation training, we propose to begin training from a randomly initialised network. In doing so, we attain comparable performance other pruning methodologies, without the need for any sophisticated pruning pipeline and stopping conditions.

\begin{table*}
    \centering
    \adjustbox{width=.8\linewidth}{
    \begin{tabular}{clcccc}
        \toprule
        
        Model & Method & Baseline Acc. (\%) & Acc. (\%) & FLOPs $\downarrow$ (\%) & Parameters $\downarrow$ (\%) \\
        \midrule
        \multirow{8}{*}{ResNet50} & G-SD-B~\cite{Liu2020RethinkingPruning} & 76.15 & 75.85 & 44 & 23 \\
        & MetaPruning~\cite{Liu2019MetaPruning:Pruning} & 76.60 & 75.40 & 50 & - \\
        & NSPPR~\cite{Zhuang2022Neuron-levelRegularizer} & 76.15 & 75.63 & 54 & - \\
        & DPFPS~\cite{Ruan2021DPFPS:Scratch} & 76.15 & 75.55 & 46 & - \\
        & S-COP~\cite{Tang2020SCOP:Pruning} & 76.15 & 75.26 & 54 & 52 \\
        & LRF-60~\cite{Joo2021LinearlyPruning} & 76.15 & 75.71 & 56 & 53 \\
        & DLRFC~\cite{He2022FilterNetworks} & 76.13 & 75.84 & 54 & 40 \\
        & \cellcolor{redx}\textbf{Ours} & \cellcolor{redx}\textbf{76.20} & \cellcolor{redx}\textbf{75.59} & \cellcolor{redx}\textbf{47} & \cellcolor{redx}\textbf{40} \\
        \bottomrule
    \end{tabular}%
    }
    \vspace{0.5em}
    \caption{\textbf{Comparison to other pruning methods on ImageNet-1K.} Our model is trained from random initialisation and with a simple magnitude based criterion. Using number of groups = 2 and with a minimum of 32 templates per layer.}
    \label{table:imagenet_results}
\end{table*}

\subsection{Ablation experiments}
\label{sec:ablation}
\paragraph{Group extension.}
\label{sec:ablation_groups}
To demonstrate the benefit of our proposed group extension, we train a VGG16 network at different pruning rates and with a varying number of groups. The results in figure \ref{fig:group_ablation} (left) show that at high pruning rates, whereby the layer will incur a large compression of features, increasing the number of groups will help. Although increasing the minimum number of templates per layer can also partially address this problem as shown in figure \ref{fig:group_ablation} (right), it would come with a much more significant computational overhead. In practice, we find that carefully selecting both the minimum number of templates and the number of groups can lead to the best performance trade-off.

\paragraph{Transformation family.}
\label{sec:transformation_family}
We explore the importance of choosing a suitable parametric family of transformations for the template filters. To do this, we first consider simple scalar multiplications of the templates, and then we consider learnable rotations. Finally, we consider the more expressive affine transformations. The results are shown in figure \ref{table:ablation_transformations}. We find that introducing more expressive transformations does improve the attainable performance, which is more significant at the higher pruning rates.

We explore the importance of selecting an appropriate parametric family of transformations for the template filters. To do this, we first consider a simple scalar multiplications applied to the templates. Subsequently, we extend our analysis to encompass learnable rotations, further expanding the range of potential transformations. Finally, to unlock the full expressive transformations, we consider the general linear group, which provide a richer and more versatile set of manipulations.

The empirical findings from our experiments are presented in the illustrative Figure \ref{table:ablation_transformations}, which serves as a visual representation of the attained results. Notably, we observe a discernible improvement in performance as we progress from simpler transformations to more expressive ones. This enhancement is particularly pronounced when operating at higher pruning rates, highlighting the significance of embracing the full spectrum of transformation possibilities.


\renewcommand{\arraystretch}{1.2}
\begin{table}
    \centering
    \adjustbox{width=.9\linewidth}{
    \begin{tabular}{ccc}
        \toprule
        Transformations & Top-1 Accuracy & Pruning Rate \\
        \hline
        Scalar & 90.62\% & 0.9 \\
        SO(3) & 92.32\% & 0.9 \\
        \textbf{GL(3)} & \textbf{92.33\%} & \textbf{0.9} \\
        \hline
        Scalar & 92.48\% & 0.7 \\
        SO(3) & 93.48\% & 0.7 \\
        \textbf{GL(3)} & \textbf{93.57\%} & \textbf{0.7} \\
        \bottomrule
    \end{tabular}%
    }
    \vspace{1em}
    \caption{\textbf{Ablating the family of transformations.} Increasing the expressivity of transformations has a small improvement in performance, suggesting that most of the network capacity is reserved for depthwise feature aggregation.}
    \label{table:ablation_transformations}
\end{table}
\renewcommand{\arraystretch}{1.}

\paragraph{Visualising learned transformations.}
\label{sec:visualisation}
Figure \ref{fig:visualising_learned_transformations} provides a comparison between the original filters, the reconstructed filters, and the pruned filters. We can discern that the the reconstructed filters are significantly distinct, thus enabling highly discriminative features for the downstream task. This result is in stark contrast with conventional pruning, which simply zeroes out these pruned filters. This visualisation highlights the significance of our novel approach which not only prunes but also actively reconstructs the filters, resulting in more informative representation of the data.

\begin{figure}[h]
\centering
\subfloat
{
    \adjustbox{width=.46\linewidth}{
    \includegraphics{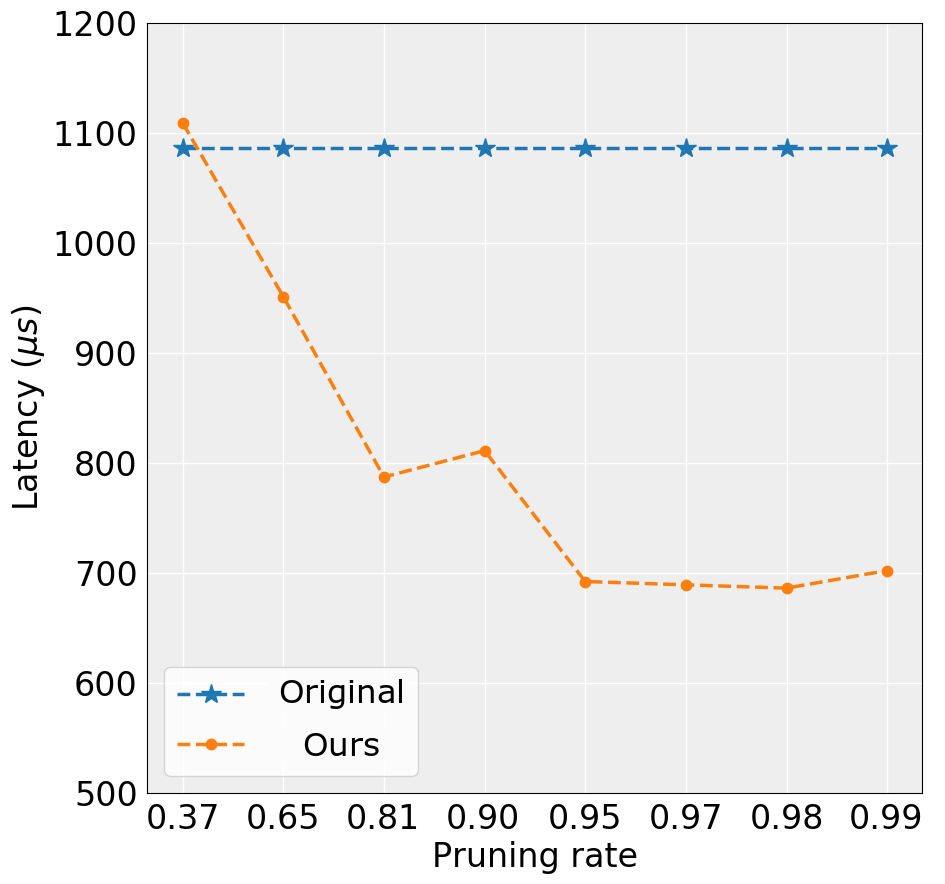}
    }
    \adjustbox{width=.365\linewidth}{
    \includegraphics{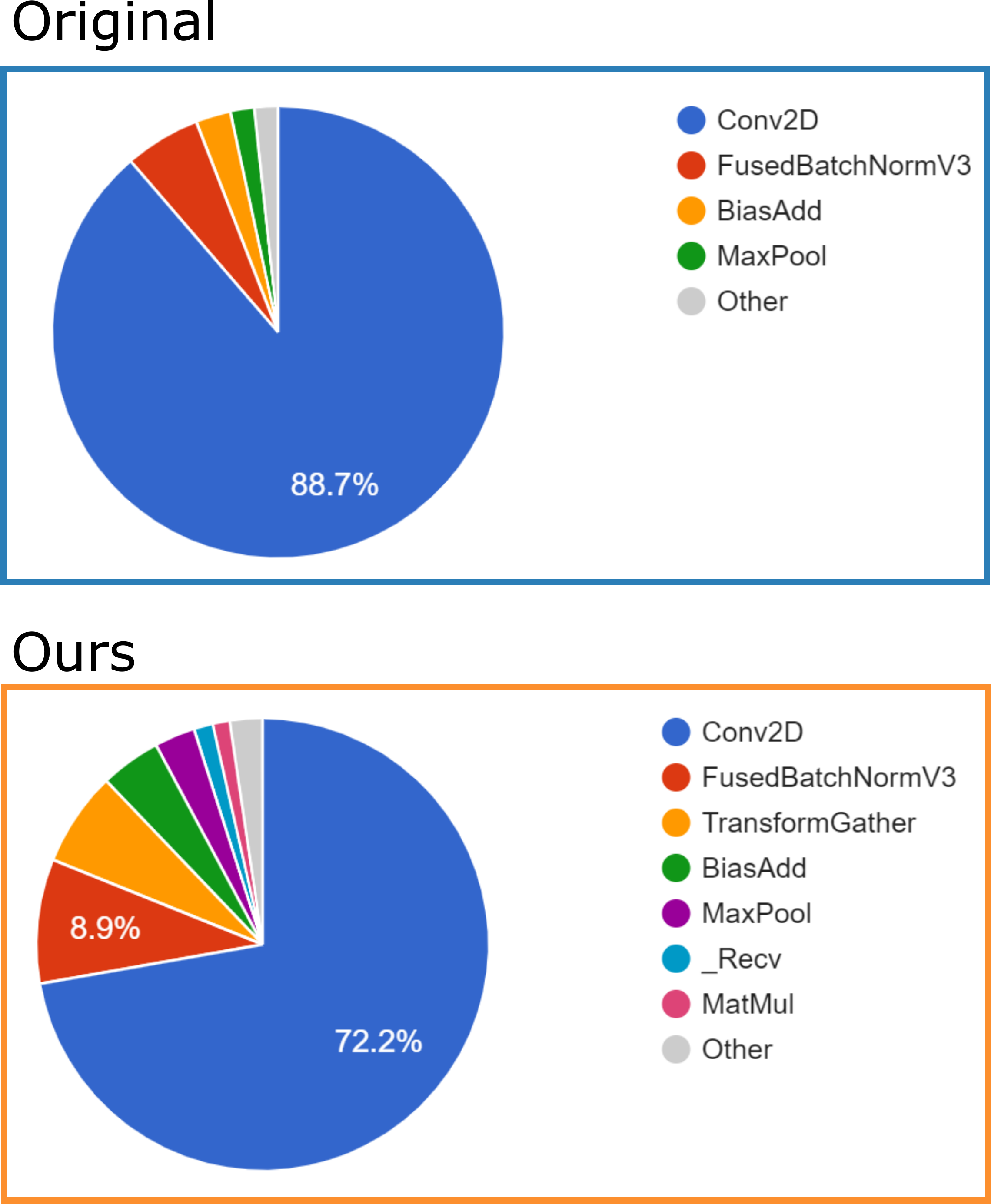}
    }
}
\caption{\textbf{Latency of the proposed decomposition with an efficient implementation.} (left) Compute time of an original VGG16 network with and without the decomposition at various pruning rates. (right) Cumulative contribution of each operation on the overall
on-device compute time.}
\label{fig:latency}
\end{figure}

\paragraph{Efficient Implementation.}
\label{sec:efficient_implementation}
To demonstrate that the theoretical reduction in FLOPs can translate to a real reduction in latency, we implement a simple CUDA kernel for the decomposed layer. The results are shown in figure \ref{fig:latency}, where we can see that at even moderate pruning rates there is a noticeable reduction in latency in comparison to the standard convolutional layer. We can also see that a large proportion of the latency is being spent on computing the template features, while a much smaller proportion comes from the scalar transformation of these features, which is implemented through a parallelized gather operation.

\section{Conclusion and Future work}
\label{sec:conclusion}
In this paper, we proposed the use of cheap transformations to reconstruct pruned filters. Instead of zeroing out the pruned filters, they are replaced with spatial transformations from the remaining set of non-pruned filters. These trained networks are able to achieve comparable or improved results on the image classification task across a range of datasets and architectures, despite using a simple magnitude based pruning criterion.
We also introduce a grouped extension that can mitigate excessive feature compression at a minimal computational cost. Our approach applied to VGG16, ResNet34 and ResNet50 is able to significantly reduce the models size and computational cost while retaining the top recognition accuracy on CIFAR-10 and ImageNet-1K datasets. 

Future research may explore potential applications in localization tasks that rely on equivariant features. Additionally, another promising direction is in data-efficient training. By incorporating hand-crafted transformations and leveraging prior knowledge of the data, it becomes possible to eliminate the necessity for the network to learn this information. Finally, we hope that this work will lead the further co-design of more weight decompositions using generalised pruning pipelines.

\printbibliography

@misc{molchanov2019importance,
      title={Importance Estimation for Neural Network Pruning}, 
      author={Pavlo Molchanov and Arun Mallya and Stephen Tyree and Iuri Frosio and Jan Kautz},
      year={2019},
      journal = {CVPR}
}

@misc{dong2017learning,
      title={Learning to Prune Deep Neural Networks via Layer-wise Optimal Brain Surgeon}, 
      author={Xin Dong and Shangyu Chen and Sinno Jialin Pan},
      year={2017},
      journal = {NeurIPS}
}

@article{Miles2023ADistillationv2,
    title = {{A closer look at the training dynamics of knowledge distillation}},
    year = {2023},
    author = {Miles, Roy and Mikolajczyk, Krystian},
    journal = {arXiv preprint}
}

@article{Yim2017ALearning,
    title = {{A Gift from Knowledge Distillation: Fast Optimization, Network Minimization and Transfer Learning}},
    year = {2017},
    journal = {CVPR},
    author = {Yim, Junho}
}

@article{Chen2018AnMobileNet,
    title = {{An Enhanced Hybrid MobileNet}},
    year = {2018},
    journal = {iCAST},
    author = {Chen, Hong Yen and Su, Chung Yen}
}

@article{Ding2019ApproximatedOptimization,
    title = {{Approximated Oracle Filter Pruning for Destructive CNN Width Optimization}},
    year = {2019},
    journal = {ICML},
    author = {Ding, Xiaohan and Ding, Guiguang and Guo, Yuchen and Han, Jungong and Yan, Chenggang},
    month = {5}
}

@article{Baalen2020BayesianPruning,
    title = {{Bayesian Bits : Unifying Quantization and Pruning}},
    year = {2020},
    author = {Baalen, Mart Van and Louizos, Christos and Nagel, Markus and Amjad, Rana Ali and Wang, Ying and Blankevoort, Tijmen and Welling, Max},
    number = {NeurIPS}
}

@inproceedings{Qin2019CAPTOR:Applications,
    title = {{CAPTOR: A Class Adaptive Filter Pruning Framework for Convolutional Neural Networks in Mobile Applications}},
    year = {2019},
    booktitle = {ASPDAC},
    author = {Qin, Zhuwei and Yu, Fuxun and Liu, Chenchen and Chen, Xiang}
}

@inproceedings{Lin2020ChannelSearch,
    title = {{Channel Pruning via Automatic Structure Search}},
    year = {2020},
    booktitle = {IJCAI},
    author = {Lin, Mingbao and Ji, Rongrong and Zhang, Yuxin and Zhang, Baochang and Wu, Yongjian and Tian, Yonghong},
    month = {1}
}

@article{Kim2015CompressionApplications,
    title = {{Compression of Deep Convolutional Neural Networks for Fast and Low Power Mobile Applications}},
    year = {2015},
    journal = {ICLR},
    author = {Kim, Yong-Deok and Park, Eunhyeok and Yoo, Sungjoo and Choi, Taelim and Yang, Lu and Shin, Dongjun}
}

@article{Miles2021CompressionApplications,
    title = {{Compression of descriptor models for mobile applications}},
    year = {2021},
    journal = {ICASSP},
    author = {Miles, Roy and Mikolajczyk, Krystian}
}

@article{Wang2019COP:Pruning,
    title = {{COP: Customized deep model compression via regularized correlation-based filter-level pruning}},
    year = {2019},
    journal = {IJCAI},
    author = {Wang, Wenxiao and Fu, Cong and Guo, Jishun and Cai, Deng and He, Xiaofei}
}

@article{Han2015DeepCoding,
    title = {{Deep Compression: Compressing Deep Neural Networks with Pruning, Trained Quantization and Huffman Coding}},
    year = {2015},
    journal = {ICLR},
    author = {Han, Song and Mao, Huizi and Dally, William J.}
}

@article{Zhuang2018Discrimination-awareNetworks,
    title = {{Discrimination-aware Channel Pruning for Deep Neural Networks}},
    year = {2018},
    journal = {NeurIPS},
    author = {Zhuang, Zhuangwei and Tan, Mingkui and Zhuang, Bohan and Liu, Jing and Guo, Yong and Wu, Qingyao and Huang, Junzhou and Zhu, Jinhui}
}

@article{Hinton2015DistillingNetwork,
    title = {{Distilling the Knowledge in a Neural Network}},
    year = {2015},
    journal = {NeurIPS},
    author = {Hinton, Geoffrey and Vinyals, Oriol and Dean, Jeff}
}

@article{Guo2020DMCP:Networks,
    title = {{DMCP: Differentiable Markov Channel Pruning for Neural Networks}},
    year = {2020},
    journal = {CVPR},
    author = {Guo, Shaopeng and Wang, Yujie and Li, Quanquan and Yan, Junjie}
}

@inproceedings{Ruan2021DPFPS:Scratch,
    title = {{DPFPS: Dynamic and Progressive Filter Pruning for Compressing Convolutional Neural Networks from Scratch}},
    year = {2021},
    booktitle = {AAAI},
    author = {Ruan, Xiaofeng and Liu, Yufan and Li, Bing and Yuan, Chunfeng and Hu, Weiming}
}

@article{Qian2020EfficientDistillation,
    title = {{Efficient Kernel Transfer in Knowledge Distillation}},
    year = {2020},
    journal = {arXiv},
    author = {Qian, Qi and Li, Hao and Hu, Juhua}
}

@article{Hayashi2019Einconv:Networks,
    title = {{Einconv: Exploring Unexplored Tensor Decompositions for Convolutional Neural Networks}},
    year = {2019},
    journal = {NeurIPS 2019},
    author = {Hayashi, Kohei and Yamaguchi, Taiki and Sugawara, Yohei and Maeda, Shin-ichi}
}

@inproceedings{He2022FilterNetworks,
    title = {{Filter Pruning via Feature Discrimination in Deep Neural Networks}},
    year = {2022},
    booktitle = {ECCV},
    author = {He, Zhiqiang and Qian, Yaguan and Wang, Yuqi and Wang, Bin and Guan, Xiaohui and Gu, Zhaoquan and Ling, Xiang and Zeng, Shaoning and Wang, Haijiang and Zhou, Wujie}
}

@article{He2018FilterAcceleration,
    title = {{Filter Pruning via Geometric Median for Deep Convolutional Neural Networks Acceleration}},
    year = {2018},
    journal = {CVPR},
    author = {He, Yang and Liu, Ping and Wang, Ziwei and Hu, Zhilan and Yang, Yi}
}

@article{Lin2019FilterPruning,
    title = {{Filter Sketch for Network Pruning}},
    year = {2019},
    journal = {arXiv},
    author = {Lin, Mingbao and Ji, Rongrong and Li, Shaojie and Ye, Qixiang and Tian, Yonghong and Liu, Jianzhuang}
}

@article{Romero2015FitNets:Nets,
    title = {{FitNets: Hints For Thin Deep Nets}},
    year = {2015},
    journal = {ICLR},
    author = {Romero, Adriana and Ballas, Nicolas and Ebrahimi Kahou, Samira and Chassang, Antoine and Gatta, Carlo and Bengio, Yoshua}
}

@article{Han2019GhostNet:Operations,
    title = {{GhostNet: More Features from Cheap Operations}},
    year = {2019},
    journal = {CVPR},
    author = {Han, Kai and Wang, Yunhe and Tian, Qi and Guo, Jianyuan and Xu, Chunjing and Xu, Chang}
}

@article{Cohen2016GroupNetworks,
    title = {{Group equivariant convolutional networks}},
    year = {2016},
    journal = {ICML},
    author = {Cohen, Taco S. and Welling, Max},
    isbn = {9781510829008}
}

@article{Li2020GroupCompression,
    title = {{Group Sparsity: The Hinge Between Filter Pruning and Decomposition for Network Compression}},
    year = {2020},
    journal = {CVPR},
    author = {Li, Yawei and Gu, Shuhang and Mayer, Christoph and Van Gool, Luc and Timofte, Radu}
}

@article{Lin2020HRank:Map,
    title = {{HRank: Filter Pruning using High-Rank Feature Map}},
    year = {2020},
    journal = {CVPR},
    author = {Lin, Mingbao and Ji, Rongrong and Wang, Yan and Zhang, Yichen and Zhang, Baochang and Tian, Yonghong and Shao, Ling}
}

@article{Krizhevsky2012ImageNetNetworks,
    title = {{ImageNet Classification with Deep Convolutional Neural Networks}},
    year = {2012},
    journal = {NeurIPS},
    author = {Krizhevsky, Alex and Sutskever, Ilya and Hinton, Geoffrey E}
}

@article{Chen2022ImprovedEnsemble,
    title = {{Improved Feature Distillation via Projector Ensemble}},
    year = {2022},
    journal = {NeurIPS},
    author = {Chen, Yudong and Wang, Sen and Liu, Jiajun and Xu, Xuwei and de Hoog, Frank and Huang, Zi}
}

@article{Miles2022InformationDistillation,
    title = {{Information Theoretic Representation Distillation}},
    year = {2022},
    journal = {BMVC},
    author = {Miles, Roy and Rodriguez, Adrian Lopez and Mikolajczyk, Krystian},
    month = {12}
}

@article{Krizhevsky2009LearningImages,
    title = {{Learning Multiple Layers of Features from Tiny Images}},
    year = {2009},
    author = {Krizhevsky, Alex}
}

@inproceedings{Joo2021LinearlyPruning,
    title = {{Linearly Replaceable Filters for Deep Network Channel Pruning}},
    year = {2021},
    booktitle = {AAAI},
    author = {Joo, Donggyu and Yi, Eojindl and Baek, Sunghyun and Kim, Junmo}
}

@inproceedings{Liu2019MetaPruning:Pruning,
    title = {{MetaPruning: Meta Learning for Automatic Neural Network Channel Pruning}},
    year = {2019},
    booktitle = {ICCV},
    author = {Liu, Zechun and Cheng, Tim Kwang-ting}
}

@article{Tan2018MnasNet:Mobile,
    title = {{MnasNet: Platform-Aware Neural Architecture Search for Mobile}},
    year = {2018},
    journal = {CVPR},
    author = {Tan, Mingxing and Chen, Bo and Pang, Ruoming and Vasudevan, Vijay and Sandler, Mark and Howard, Andrew and Le, Quoc V.}
}

@article{Fox2018MobileNetV2:Bottlenecks,
    title = {{MobileNetV2: Inverted Residuals and Linear Bottlenecks}},
    year = {2018},
    journal = {CVPR},
    author = {Fox, Michael H. and Kim, Kyungmee and Ehrenkrantz, David}
}

@article{Miles2023MobileVOS:Distillation,
    title = {{MobileVOS: Real-Time Video Object Segmentation Contrastive Learning meets Knowledge Distillation}},
    year = {2023},
    journal = {CVPR},
    author = {Miles, Roy and Yucel, Mehmet Kerim and Manganelli, Bruno and Saa-Garriga, Albert},
    month = {3}
}

@article{Subramaniam2020N2NSkip:Connections,
    title = {{N2NSkip: Learning Highly Sparse Networks using Neuron-to-Neuron Skip Connections}},
    year = {2020},
    journal = {BMVC},
    author = {Subramaniam, Arvind and Sharma, Avinash}
}

@article{Hu2016NetworkArchitectures,
    title = {{Network Trimming: A Data-Driven Neuron Pruning Approach towards Efficient Deep Architectures}},
    year = {2016},
    journal = {arXiv},
    author = {Hu, Hengyuan and Peng, Rui and Tai, Yu-Wing and Tang, Chi-Keung}
}

@inproceedings{Zhuang2022Neuron-levelRegularizer,
    title = {{Neuron-level Structured Pruning using Polarization Regularizer}},
    year = {2022},
    booktitle = {NeurIPS},
    author = {Zhuang, Tao and Zhang, Zhixuan and Huang, Yuheng and Zeng, Xiaoyi and Shuang, Kai and Li, Xiang}
}

@article{Yu2018NISP:Propagation,
    title = {{NISP: Pruning Networks Using Neuron Importance Score Propagation}},
    year = {2018},
    journal = {CVPR},
    author = {Yu, Ruichi and Li, Ang and Chen, Chun Fu and Lai, Jui Hsin and Morariu, Vlad I. and Han, Xintong and Gao, Mingfei and Lin, Ching Yung and Davis, Larry S.}
}

@article{Lecun1990OptimalDamage,
    title = {{Optimal Brain Damage}},
    year = {1990},
    journal = {NeurIPS},
    author = {Lecun, Yann}
}

@article{Li2017PruningConvnets,
    title = {{Pruning Filters For Efficient Convnets}},
    year = {2017},
    journal = {ICLR},
    author = {Li, Hao and Kadav, Asim and Durdanovic, Igor and Samet, Hanan and Graf, Hans Peter},
    arxivId = {1608.08710v3}
}

@inproceedings{Liu2020RethinkingPruning,
    title = {{Rethinking Class-Discrimination Based CNN Channel Pruning}},
    year = {2020},
    booktitle = {arXiv preprint},
    author = {Liu, Yuchen and Wentzlaff, David and Kung, S. Y.},
    month = {4}
}

@inproceedings{Tang2020SCOP:Pruning,
    title = {{SCOP: Scientific Control for Reliable Neural Network Pruning}},
    year = {2020},
    booktitle = {NeurIPS},
    author = {Tang, Yehui and Wang, Yunhe and Xu, Yixing and Tao, Dacheng and Xu, Chunjing and Xu, Chao and Xu, Chang},
    month = {10}
}

@article{Howard2019SearchingMobileNetV3,
    title = {{Searching for MobileNetV3}},
    year = {2019},
    journal = {ICCV},
    author = {Howard, Andrew and Sandler, Mark and Chu, Grace and Chen, Liang-Chieh and Chen, Bo and Tan, Mingxing and Wang, Weijun and Zhu, Yukun and Pang, Ruoming and Vasudevan, Vijay and Le, Quoc V. and Adam, Hartwig}
}

@article{Hassibi1993SecondSurgeon,
    title = {{Second order derivatives for network pruning: Optimal Brain Surgeon}},
    year = {1993},
    journal = {NeurIPS},
    author = {Hassibi, Babak and Stork, David G}
}

@article{Zhang2018ShuffleNet:Devices,
    title = {{ShuffleNet: An Extremely Efficient Convolutional Neural Network for Mobile Devices}},
    year = {2018},
    journal = {CVPR},
    author = {Zhang, Xiangyu and Zhou, Xinyu and Lin, Mengxiao},
    arxivId = {1707.01083v2}
}

@article{Yu2018SlimmableNetworks,
    title = {{Slimmable Neural Networks}},
    year = {2018},
    journal = {ICLR},
    author = {Yu, Jiahui and Yang, Linjie and Xu, Ning and Yang, Jianchao and Huang, Thomas}
}

@inproceedings{Lee2019SnIP:Sensitivity,
    title = {{SnIP: Single-shot network pruning based on connection sensitivity}},
    year = {2019},
    booktitle = {ICLR},
    author = {Lee, Namhoon and Ajanthan, Thalaiyasingam and Torr, Philip H.S.},
    arxivId = {1810.02340}
}

@article{Tucker1966SomeAnalysis,
    title = {{Some mathematical notes on three-mode factor analysis}},
    year = {1966},
    journal = {Psychometrika},
    author = {Tucker, Ledyard R}
}

@inproceedings{Jaderberg2014SpeedingExpansions,
    title = {{Speeding up convolutional neural networks with low rank expansions}},
    year = {2014},
    booktitle = {BMVC},
    author = {Jaderberg, Max and Vedaldi, Andrea and Zisserman, Andrew}
}

@article{Detone2018SuperPoint:Description,
    title = {{SuperPoint: Self-supervised interest point detection and description}},
    year = {2018},
    journal = {ICPR},
    author = {Detone, Daniel and Malisiewicz, Tomasz and Rabinovich, Andrew}
}

@article{Hitchcock2015TheProducts,
    title = {{The Expression of a Tensor or a Polyadic as a Sum of Products}},
    year = {2015},
    journal = {JMP},
    author = {Hitchcock, Frank L.}
}

@inproceedings{Luo2017ThiNet:Compression,
    title = {{ThiNet: A Filter Level Pruning Method for Deep Neural Network Compression}},
    year = {2017},
    booktitle = {ICCV},
    author = {Luo, Jian Hao and Wu, Jianxin and Lin, Weiyao}
}

@article{Yu2019UniversallyTechniques,
    title = {{Universally Slimmable Networks and Improved Training Techniques}},
    year = {2019},
    journal = {ICCV},
    author = {Yu, Jiahui and Huang, Thomas}
}

@article{Zhao2019VariationalPruning,
    title = {{Variational Convolutional Neural Network Pruning}},
    year = {2019},
    journal = {CVPR},
    author = {Zhao, Chenglong and Ni, Bingbing and Zhang, Jian and Zhao, Qiwei}
}

@article{Simonyan2015VeryRecognition,
    title = {{Very Deep Convolutional Networks For Large-scale Image Recognition}},
    year = {2015},
    journal = {ICLR},
    author = {Simonyan, Karen and Zisserman, Andrew}
}

@article{Wang2018WideNets,
    title = {{Wide Compression : Tensor Ring Nets}},
    year = {2018},
    journal = {CVPR},
    author = {Wang, Wenqi and Eriksson, Brian and Wang, Wenlin}
}

@article{Chollet2017Xception:Convolutions,
    title = {{Xception: Deep Learning with Depthwise Separable Convolutions}},
    year = {2017},
    journal = {CVPR},
    author = {Chollet, François},
    arxivId = {1610.02357v3}
}

\end{document}